\documentclass[10pt, a4paper]{article}
\usepackage{float}

\usepackage[final]{lrec2026}
\usepackage{booktabs}
\usepackage{multirow}
\usepackage{url}
\graphicspath{{images/}}

\title{Mining Large Language Models for Low-Resource Language Data: Comparing Elicitation Strategies for Hausa and Fongbe}

\name{Mahounan Pericles Adjovi, Roald Eiselen, Prasenjit Mitra}

\name{Mahounan Pericles Adjovi$^1$, Roald Eiselen$^2$, Prasenjit Mitra$^1$}

\address{
    $^1$Carnegie Mellon University Africa, Kigali, Rwanda \\
    $^2$Centre for Text Technology, North-West University, Potchefstroom, South Africa \\
    madjovi@andrew.cmu.edu, Roald.Eiselen@nwu.ac.za, prasenjm@andrew.cmu.edu
}

\abstract{
Large language models (LLMs) are trained on data contributed by low-resource language communities, including curated datasets such as MasakhaNER and MAFAND-MT, yet the linguistic knowledge encoded in these models remains accessible only through commercial APIs. This paper investigates whether strategic prompting can extract usable text data from LLMs for two West African languages: Hausa (Afroasiatic, approximately 80 million speakers) and Fongbe (Niger-Congo, approximately 2 million speakers). We systematically compare six elicitation task types: creative writing, functional text, structured knowledge, dialogue, topic-switching probes, and constrained generation across two commercial LLMs (GPT-4o Mini and Gemini 2.5 Flash). Generated outputs are evaluated on linguistic accuracy, lexical diversity, domain coverage, and code-switching rates through automatic assessment metrics. Our findings reveal that elicitation strategy significantly affects output quality and that optimal strategies differ by language: Hausa benefits from volume-maximizing tasks such as functional text and dialogue, while Fongbe requires constraint-heavy prompts that enforce monolingual output. GPT-4o Mini extracts 6--41$\times$ more usable target-language words per API call than Gemini, though Gemini achieves higher language purity for Fongbe on constrained tasks. We provide a practical framework for low-resource language communities to maximize usable data extraction from LLMs and release all generated corpora and code.
\\ \newline }

\begin{document}

\maketitleabstract

\section{Introduction}

Natural Language Processing (NLP) technologies remain largely inaccessible to speakers of most African languages due to severe data scarcity \citep{Joshi-etal-2020}. Languages such as Fongbe, a national language of Benin, and Hausa, widely spoken across West Africa, suffer from limited digital text resources despite having millions of speakers. Meanwhile, large language models (LLMs) have been trained on web-scale data that includes contributions from these language communities, including curated academic datasets such as MasakhaNER 2.0 \citep{Adelani-etal-2022-masakhaner} and MAFAND-MT \citep{Adelani-etal-2022-mafand}. The linguistic knowledge absorbed from these sources resides within commercial LLMs, yet it flows back to these communities only through paid API access. A natural question arises: can we systematically extract usable language data from these models to create new resources for the very communities whose data helped build them?

This question has both practical and ethical significance. On the practical side, low-resource language communities face a critical bootstrapping problem: building NLP systems requires data, but data collection is expensive and slow. If LLMs can serve as an efficient source of text data across diverse domains, this could accelerate resource creation for languages where text expansion remains a critical priority gap. On the ethical side, the relationship between LLM training data and community benefit is asymmetric: language communities contribute data that increases the commercial value of LLMs, yet receive limited benefit in return. Developing systematic methods for extracting linguistic knowledge from LLMs represents a practical step toward rebalancing this relationship.

Extracting usable language data from LLMs is non-trivial for several reasons. First, LLM generation quality varies dramatically across low-resource languages, with substantial performance gaps documented even among African languages with millions of speakers \citep{Robinson-etal-2023,Hendy-etal-2023}. Second, generated text frequently exhibits code-switching with colonial languages: English for Hausa, French for Fongbe, reducing its utility as monolingual training data \citep{Orife-etal-2020}. Third, for tonal languages like Fongbe, LLMs frequently produce missing or incorrect diacritics, which are obligatory in standard orthography and distinguish lexical meaning \citep{Lefebvre-Brousseau-2002}. Fourth, it is unclear which prompting strategies maximize both the quantity and quality of extractable data.


Previous work has examined LLMs for low-resource language tasks primarily through machine translation \citep{Robinson-etal-2023,Hendy-etal-2023} or data augmentation for specific downstream tasks \citep{Whitehouse-etal-2023}. The Fikira dataset \citep{Adelani-etal-2024-fikira} demonstrated that instruction-tuned models can generate reasoning data for African languages, but did not compare across elicitation task types. To our knowledge, no study has systematically explored elicitation strategies to assess which tasks may yield the most usable data per API call for low-resource African languages. This work presents an early exploratory investigation into this question, with the goal of identifying promising directions rather than drawing definitive conclusions.

\subsection{Research Question}

We address the following central research question:

\begin{quote}
\textit{Which types of elicitation tasks maximize the quantity and quality of usable text data that can be extracted from large language models for Hausa and Fongbe?}
\end{quote}

This is operationalized through three sub-questions:

\begin{enumerate}
    \item How does the linguistic quality of LLM-generated text vary across elicitation task types for Hausa and Fongbe?
    \item Which elicitation strategies produce the greatest lexical diversity and domain coverage per API call?
    \item Do optimal elicitation strategies differ between languages with different levels of LLM support?
\end{enumerate}


\subsection{Summary of Contributions}

\begin{itemize}
    \item A systematic taxonomy of LLM elicitation strategies for low-resource language data extraction, evaluated across six task types for two typologically distinct West African languages (Section~\ref{sec:methodology}).
    \item An empirical comparison of two commercial LLMs revealing that GPT-4o Mini generates 6--41 times more usable target-language text than Gemini 2.5 Flash, with language-specific optimal strategies (Section~\ref{sec:results}).
    \item A practical framework and released corpora enabling low-resource language communities to replicate our methodology (Section~\ref{sec:discussion}).
\end{itemize}

\section{Related Work}
\label{sec:related}

\subsection{African Language NLP Resources}

Research on African language technology has accelerated significantly since 2019. The Masakhane project established a participatory approach to machine translation across more than 30 African languages \citep{Nekoto-etal-2020}. Subsequent efforts produced standardized benchmarks: MasakhaNER 2.0 for NER across 20 languages \citep{Adelani-etal-2022-masakhaner}, MasakhaPOS for part-of-speech tagging \citep{Adelani-etal-2023-masakhpos}, and MAFAND-MT for news-domain machine translation \citep{Adelani-etal-2022-mafand}.

Despite these advances, resource availability remains severely unbalanced. Under the taxonomy of \citet{Joshi-etal-2020}, Hausa falls in mid-tiers (3--4) given its international media presence, while Fongbe falls closer to the lowest tiers (0--1). Continental surveys confirm that most African languages lack sufficient corpora \citep{Hedderich-etal-2021}. Our work proposes LLM-based data extraction as a scalable complement to manual corpus construction.

\subsection{LLMs for Low-Resource Languages}

\citet{Robinson-etal-2023} showed that ChatGPT degrades significantly for low-resource African languages. \citet{Hendy-etal-2023} found systematic quality drops for languages with limited web presence. AfriDoc-MT \citep{Alabi-etal-2025-afridocmt} evaluated document-level translation for African languages including Hausa. African-centric models such as AfriBERTa \citep{Ogueji-etal-2021} and AfroXLMR \citep{Alabi-etal-2022} outperform multilingual baselines but focus on comprehension rather than generation.

A critical gap persists: none of these studies investigate which \textit{types} of prompts maximize extractable language data. Our work reframes the question from ``how well do LLMs translate into language X?'' to ``which prompting strategies extract the most usable data from LLMs for language X?''

\subsection{Data Augmentation and Synthetic Data}

Data augmentation encompasses Easy Data Augmentation (EDA) \citep{Wei-Zou-2019}, back-translation \citep{Sennrich-etal-2016}, and LLM-based generation \citep{Schick-Schuetze-2021}. \citet{Whitehouse-etal-2023} found mixed results for low-resource LLM augmentation. \citet{Dai-Adel-2020} showed augmentation effectiveness depends on method and dataset size. The Fikira dataset \citep{Adelani-etal-2024-fikira} generated reasoning data for African languages but did not compare elicitation strategies. These studies evaluate augmentation for specific downstream tasks rather than investigating which strategies maximize general corpus utility.

\section{Methodology}
\label{sec:methodology}


We design a controlled experiment comparing six elicitation task types across two LLMs and two languages. All prompts, scripts, and evaluation code are released publicly (see ~\ref{app:repo}). Full prompt structures and examples are provided in ~\ref{app:prompts}.

\subsection{Elicitation Task Taxonomy}


Table~\ref{tab:task_taxonomy} summarizes six task types, each probing a different dimension of LLM linguistic knowledge (see ~\ref{app:prompts} for full prompt details).

\begin{table}[H]
\centering
\footnotesize
\setlength{\tabcolsep}{3pt}
\renewcommand{\arraystretch}{0.85}
\begin{tabularx}{\columnwidth}{p{1.7cm}Xc}
\toprule
\textbf{Task Type} & \textbf{Rationale} & \textbf{N} \\
\midrule
Creative Writing {\scriptsize (poems, stories, folktales, songs, proverbs)} & Tests deep cultural and linguistic knowledge; generates domain-diverse narrative text & 25 \\
Functional Text {\scriptsize (letters, instructions, news, recipes, announcements)} & Tests practical domain coverage; generates text useful for downstream NLP & 25 \\
Structured Knowledge {\scriptsize (definitions, grammar examples, vocabulary lists, translations)} & Tests metalinguistic knowledge; produces high-density lexical output & 25 \\
Dialogue {\scriptsize (conversations, interviews, negotiations, family discussions)} & Tests colloquial register and spoken-form generation & 25 \\
Topic Switching {\scriptsize (domestic{\textrightarrow}sports, narrative shifts, knowledge switches)} & Tests language maintenance robustness under topic changes & 25 \\
Constrained Gen. {\scriptsize (vocab-constrained, no-code-switching, technical monolingual)} & Tests ability to stay in target language under explicit constraints & 25 \\
\bottomrule
\end{tabularx}
\caption{Elicitation task taxonomy: 6 types $\times$ 25 prompts = 150 per language.}
\label{tab:task_taxonomy}
\end{table}


\textbf{Creative Writing} prompts request poems, folktales, stories, songs, and proverbs about culturally relevant themes. \textbf{Functional Text} prompts request letters, instructions, news articles, recipes, and announcements. \textbf{Structured Knowledge} prompts request definitions, cultural explanations, grammar examples, vocabulary lists, and translations. \textbf{Dialogue} prompts request conversations in varied social contexts (market, clinic, family, interview, negotiation). \textbf{Topic Switching} prompts begin on a familiar topic and switch to a domain typically discussed in colonial languages, requiring continuation in the target language. \textbf{Constrained Generation} prompts impose vocabulary constraints, no-code-switching rules, length requirements, and structural formats.

\subsection{Languages}

\textbf{Hausa} (ISO 639-3: hau) is an Afroasiatic language spoken by approximately 80--100 million people across Nigeria, Niger, and neighboring countries. It features grammatical gender, rich morphology, and complex tense-aspect-mood marking \citep{Newman-2000}. It has a standardized Latin orthography, substantial web presence, and is included in XLM-RoBERTa \citep{Conneau-etal-2020}. The colonial contact language is English.

\textbf{Fongbe} (ISO 639-3: fon) is a Niger-Congo Gbe language spoken by approximately 2 million people in Benin. It features serial verb constructions and a three-tone system with obligatory diacritic marking \citep{Lefebvre-Brousseau-2002}. Tone distinguishes meaning: \textit{k\'{o}} (high) = ``harvest,'' \textit{k\`{o}} (low) = ``build,'' \textit{k\^{o}} (falling) = ``neck.'' Fongbe has minimal web presence and is absent from XLM-RoBERTa. The colonial contact language is French.

\subsection{Models and Prompt Design}
\label{sec:prompt-design}

We evaluate GPT-4o Mini (OpenAI) and Gemini 2.5 Flash (Google), both accessed with temperature 0.7, top-p 0.95, max output 1,024 tokens, and a system prompt requiring target-language output. Each prompt template contains placeholders (\texttt{\{language\}}, \texttt{\{language\_culture\}}, \texttt{\{colonial\_language\}}) substituted at generation time.

All prompts are written in English. This choice reflects a deliberate experimental design decision: English-language prompting provides a controlled, reproducible interface that does not require annotators to be proficient in Hausa or Fongbe, and enables direct comparability across languages. We acknowledge that prompting directly in the target language may yield different results and we consider this a promising direction for future work (Section~\ref{sec:future}). Preliminary evidence from related work suggests that target-language prompting can improve output quality for well-resourced languages, though its effects for very low-resource languages like Fongbe remain untested.

Each prompt is sent once per model per language: $150 \times 2 \times 2 = 600$ API calls. Outputs are saved as JSON with resumability support.




\subsection{Evaluation Framework}

We evaluate outputs using: \textbf{Output Validity} (minimum 20 tokens); \textbf{Lexical Diversity} (TTR, hapax ratio, vocabulary size); \textbf{Language Fidelity} via GlotLID \citep{Kargaran-etal-2023}, a fastText classifier covering 2,000+ languages including \texttt{hau\_Latn} and \texttt{fon\_Latn}, applied at document and sentence levels; \textbf{Diacritic Analysis} for Fongbe (tonal vowel ratio); \textbf{Repetition Detection} (4-gram and sentence repetition); and \textbf{Reference Overlap} (character trigram cosine similarity against MasakhaNER 2.0 training text for Hausa).

\section{Results}
\label{sec:results}

We report results from 600 API calls (150 prompts $\times$ 2 models $\times$ 2 languages).

\subsection{Output Validity}

Table~\ref{tab:validity} reports the percentage of outputs exceeding the 20-token minimum and the average word count per condition.

\begin{table}[H]
\centering
\small
\setlength{\tabcolsep}{3pt}
\begin{tabularx}{\columnwidth}{lXXXX}
\toprule
& \multicolumn{2}{c}{\textbf{Gemini}} & \multicolumn{2}{c}{\textbf{GPT-4o Mini}} \\
\cmidrule(lr){2-3} \cmidrule(lr){4-5}
\textbf{Task} & \textbf{Fon} & \textbf{Hau} & \textbf{Fon} & \textbf{Hau} \\
\midrule
Creative       & 28/18  & 76/27  & 100/90  & 100/104 \\
Functional     & 36/17  & 68/20  & 100/153 & 100/205 \\
Structured     & 40/18  & 56/20  & 100/92  & 100/114 \\
Dialogue       & 80/23  & 52/20  & 100/145 & 100/190 \\
Topic Switch   &  4/15  & 88/73  & 100/157 & 100/190 \\
Constrained    & 20/17  & 92/34  & 100/78  & 100/125 \\
\midrule
\textbf{Overall} & \textbf{35/18} & \textbf{72/32} & \textbf{100/119} & \textbf{100/155} \\
\bottomrule
\end{tabularx}
\caption{Output validity (\% valid/avg.\ words) by task and model.}
\label{tab:validity}
\end{table}


GPT-4o Mini achieves 100\% validity across all 12 conditions, producing outputs averaging 119 words for Fongbe and 155 for Hausa. Gemini generates much shorter responses: 18 words on average for Fongbe (35\% valid) and 32 for Hausa (72\% valid). The disparity is most extreme for Fongbe topic-switching, where Gemini produces valid output for only 4\% of prompts.

\subsection{Language Fidelity}

Table~\ref{tab:fidelity} reports document-level target language detection using GlotLID.

\begin{table}[!ht]
\centering
\small
\begin{tabular}{lcccc}
\toprule
& \multicolumn{2}{c}{\textbf{Gemini}} & \multicolumn{2}{c}{\textbf{GPT-4o Mini}} \\
\cmidrule(lr){2-3} \cmidrule(lr){4-5}
\textbf{Task} & \textbf{Fon} & \textbf{Hau} & \textbf{Fon} & \textbf{Hau} \\
\midrule
Creative       & 60  & 92  & 48  & 100 \\
Functional     & 68  & 96  & 40  & 100 \\
Structured     & 40  & 72  & 52  & 100 \\
Dialogue       & 12  & 76  & 60  & 100 \\
Topic Switch   & 96  & 100 & 32  & 100 \\
Constrained    & 100 & 100 & 88  & 100 \\
\midrule
\textbf{Overall} & \textbf{63} & \textbf{89} & \textbf{53} & \textbf{100} \\
\bottomrule
\end{tabular}
\caption{Document-level target language detection (\%) by GlotLID, per task and model.}
\label{tab:fidelity}
\end{table}

Hausa outputs are reliably identified: 89\% for Gemini and 100\% for GPT-4o Mini. Fongbe shows greater variation. Gemini achieves its highest Fongbe fidelity on constrained generation (100\%) and topic switching (96\%), while GPT-4o Mini scores highest on constrained generation (88\%) but poorly on topic switching (32\%). When Fongbe outputs are misidentified, GlotLID most frequently labels them as English (32 cases), Yoruba (24), or French (23), suggesting code-switching contamination or generation in related Gbe languages.

At the sentence level, constrained generation achieves the lowest code-switching rates across both models (0.01--0.09). GPT-4o Mini shows consistently low code-switching for Hausa (0.01--0.16) but elevated rates for Fongbe (0.25--0.66), indicating frequent interspersion of French or English sentences within otherwise Fongbe text.

\subsection{Lexical Diversity}

Table~\ref{tab:diversity} reports TTR and vocabulary size.

\begin{table}[!ht]
\centering
\small
\begin{tabular}{lcccc}
\toprule
& \multicolumn{2}{c}{\textbf{Gemini}} & \multicolumn{2}{c}{\textbf{GPT-4o Mini}} \\
\cmidrule(lr){2-3} \cmidrule(lr){4-5}
\textbf{Task} & \textbf{Fon} & \textbf{Hau} & \textbf{Fon} & \textbf{Hau} \\
\midrule
\multicolumn{5}{l}{\textit{Type-Token Ratio}} \\
Creative       & .93 & .92 & .58 & .67 \\
Functional     & .88 & .92 & .48 & .60 \\
Structured     & .96 & .95 & .60 & .71 \\
Dialogue       & .88 & .92 & .46 & .58 \\
Topic Switch   & .89 & .81 & .48 & .63 \\
Constrained    & .89 & .82 & .54 & .67 \\
\midrule
\multicolumn{5}{l}{\textit{Avg.\ Vocabulary Size}} \\
Creative       & 16 & 24 & 50  & 68 \\
Functional     & 15 & 19 & 74  & 117 \\
Structured     & 18 & 19 & 48  & 73 \\
Dialogue       & 20 & 18 & 66  & 108 \\
Topic Switch   & 13 & 53 & 70  & 117 \\
Constrained    & 15 & 27 & 32  & 74 \\
\bottomrule
\end{tabular}
\caption{Lexical diversity measured by Type-Token Ratio (TTR) and average vocabulary size per condition.}
\label{tab:diversity}
\end{table}

Gemini's higher TTR (0.81--0.96 vs.\ 0.46--0.71) is largely an artifact of output length. In absolute terms, GPT-4o Mini yields 13,895 unique Hausa and 8,478 Fongbe word tokens across all outputs, versus Gemini's 3,977 and 2,427---a 3.5$\times$ advantage.

\subsection{Fongbe Diacritic Analysis}

GPT-4o Mini produces diacritics in 96--100\% of Fongbe outputs, with diacritic-to-alphabetic ratios of 0.24--0.37. Gemini is less reliable: only 36\% of dialogue outputs contain diacritics (ratio 0.02), versus 100\% for constrained generation (ratio 0.31). Explicit constraints help Gemini activate Fongbe orthographic knowledge that unconstrained tasks fail to elicit.

\subsection{Extraction Efficiency}

\begin{figure}[H]
\centering
\includegraphics[width=0.45\textwidth]{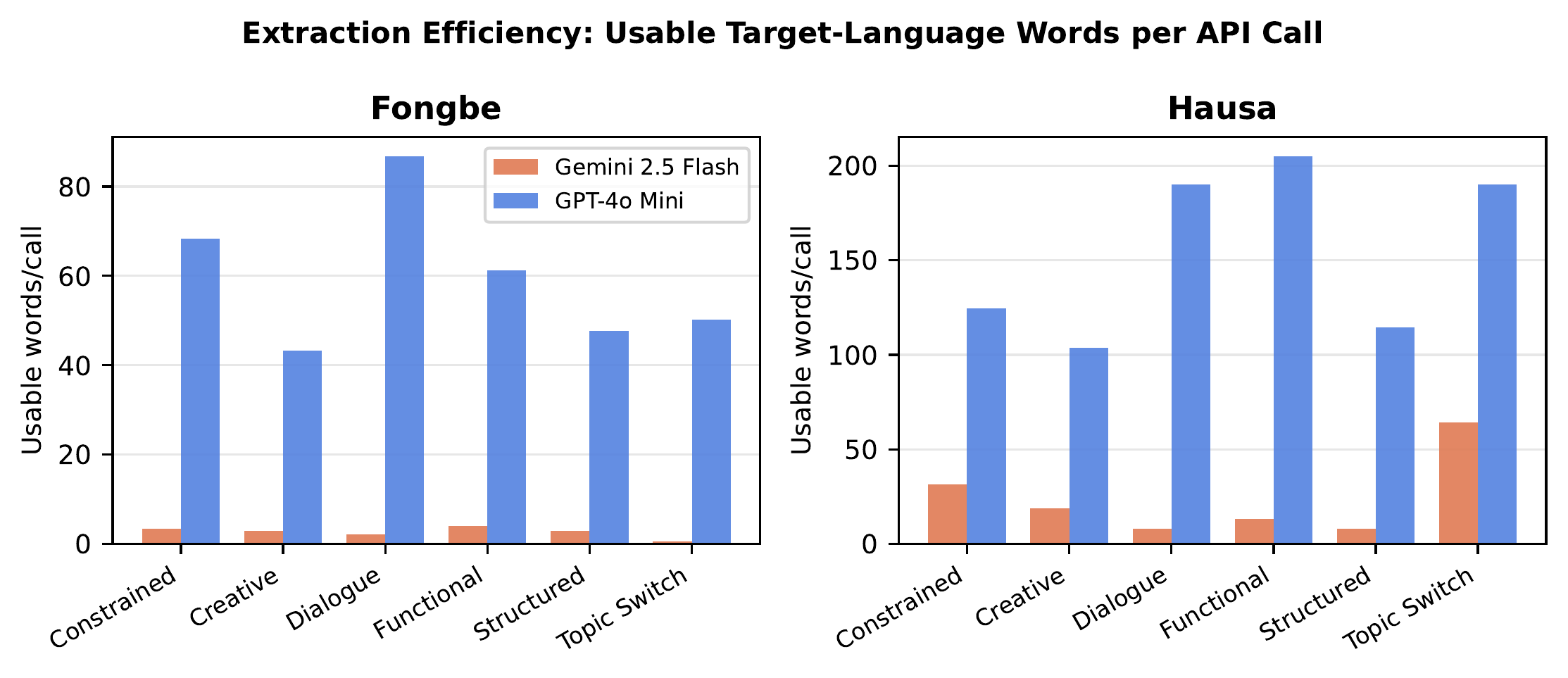}
\caption{Extraction efficiency: usable target-language words per API call, by model, language, and task type. GPT-4o Mini dominates across all conditions; the gap is most extreme for Fongbe.}
\label{fig:efficiency}
\end{figure}

Table~\ref{tab:efficiency} presents usable words per API call (from outputs that are both valid and detected as the target language). Figure~\ref{fig:efficiency} (~\ref{app:figures}) visualizes these differences.

\begin{table}[H]
\centering
\small
\begin{tabular}{llrr}
\toprule
\textbf{Model} & \textbf{Task} & \textbf{Fon} & \textbf{Hau} \\
\midrule
\multirow{7}{*}{Gemini} & Creative & 0.0 & 20.7 \\
 & Functional & 1.7 & 13.5 \\
 & Structured & 0.9 & 5.8 \\
 & Dialogue & 0.0 & 7.0 \\
 & Topic Switch & 0.0 & 70.6 \\
 & Constrained & 5.7 & 32.7 \\
 & \textbf{Per call} & \textbf{1.4} & \textbf{25.0} \\
\midrule
\multirow{7}{*}{GPT-4o} & Creative & 37.5 & 103.7 \\
 & Functional & 55.0 & 204.8 \\
 & Structured & 49.4 & 114.5 \\
 & Dialogue & 80.8 & 190.0 \\
 & Topic Switch & 50.6 & 190.1 \\
 & Constrained & 69.6 & 124.6 \\
 & \textbf{Per call} & \textbf{57.2} & \textbf{154.6} \\
\bottomrule
\end{tabular}
\caption{Usable target-language words per API call. GPT-4o Mini is 6$\times$ more efficient for Hausa, 41$\times$ for Fongbe.}
\label{tab:efficiency}
\end{table}

GPT-4o Mini extracts 154.6 usable Hausa words per call (6$\times$ Gemini) and 57.2 Fongbe words (41$\times$ Gemini). The most efficient strategies differ by language: for Hausa, functional text and dialogue maximize extraction (190--205 words/call); for Fongbe, dialogue (80.8) and constrained generation (69.6) are most productive. Gemini's Fongbe extraction is near zero for most tasks.

\begin{figure}[H]
\centering
\includegraphics[width=0.45\textwidth]{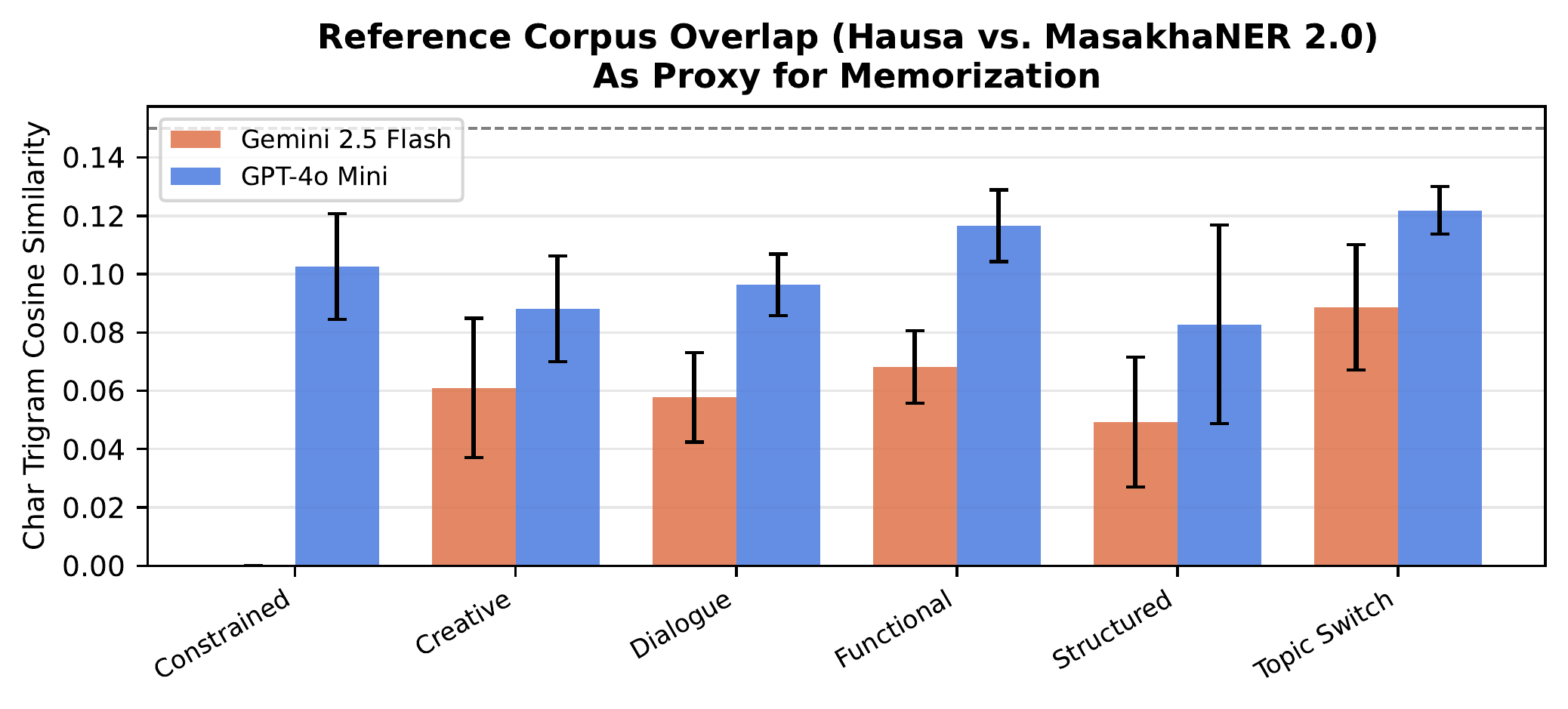}
\caption{Character trigram cosine similarity between generated Hausa text and MasakhaNER~2.0 training text, used as a proxy for potential memorization. All values are well below 0.15 (dashed line), suggesting outputs represent novel generation rather than training data reproduction.}
\label{fig:overlap}
\end{figure}

Repetition rates are low across all conditions ($<$0.06). Reference corpus overlap shows GPT-4o Mini's Hausa outputs have higher character trigram similarity to MasakhaNER~2.0 text (cosine 0.10 vs.\ 0.07 for Gemini). Crucially, both values are well below 0.15, a conservative threshold above which near-verbatim reproduction would become plausible. The elevated similarity for GPT-4o Mini most likely reflects that this model generates more natural Hausa text whose statistical profile resembles existing Hausa corpora---an indicator of generation quality rather than memorization. All cosine values by task type are visualized in Figure~\ref{fig:overlap} (\ref{app:figures}).

\section{Discussion}
\label{sec:discussion}

\subsection{Optimal Elicitation Strategies by Language}

Our results confirm that optimal strategies differ substantially between languages (RQ3).

For \textbf{Hausa}, functional text and dialogue yield the most usable words (190--205 per call with GPT-4o Mini), while constrained generation and topic switching achieve the highest language fidelity (100\% for both models). Hausa is sufficiently represented in LLM training data to sustain generation across diverse task types.

For \textbf{Fongbe}, constrained generation emerges as the most reliable strategy: highest language fidelity (100\% Gemini, 88\% GPT-4o Mini), best diacritic ratios, and lowest code-switching. Communities working with extremely low-resource languages should prioritize constrained generation prompts that explicitly require monolingual output and specify target-language vocabulary.

The Hausa--Fongbe gap is consistently large: GPT-4o Mini achieves 100\% vs.\ 53\% language fidelity, produces 2.7$\times$ more usable words per call, and exhibits 4--10$\times$ lower code-switching rates. This disparity likely reflects training data representation rather than inherent linguistic difficulty.

\subsection{Training Data as a Confounding Factor}

The performance gap between GPT-4o Mini and Gemini 2.5 Flash and between Hausa and Fongbe most plausibly reflects differences in training data composition rather than architectural differences per se. \citet{Robinson-etal-2023} demonstrated that ChatGPT performance degrades sharply for languages underrepresented in web-crawled pretraining data, and \citet{Hendy-etal-2023} showed systematic quality drops correlate with web presence rather than linguistic complexity. Hausa has a substantial international media presence (BBC Hausa, VOA Hausa), whereas Fongbe has minimal digital footprint. If Gemini's training data includes proportionally less Hausa and Fongbe text than GPT-4o Mini's, this would fully explain the extraction efficiency gap without invoking any architectural cause. Unfortunately, neither OpenAI nor Google discloses the language-level composition of their training data, making this hypothesis untestable with current information. Future work using open-weight models with documented training corpora (e.g., BLOOM, Llama variants) could help disentangle data from architecture effects.

\subsection{Implications for Resource Creation}

Our findings yield practical recommendations. First, \textbf{model selection matters more than task selection}: switching from Gemini to GPT-4o Mini increases Fongbe efficiency by 41$\times$, whereas task variation within GPT-4o Mini yields only 2$\times$ difference. Second, \textbf{explicit constraints improve fidelity}: constrained generation consistently achieves the highest language purity and diacritic accuracy. Third, \textbf{post-hoc filtering is essential}: even the best Fongbe condition produces 12\% non-target outputs; GlotLID filtering can remove contaminated text. Fourth, \textbf{cost-efficiency is compelling}: GPT-4o Mini extracted 23,192 usable Hausa words and 8,574 Fongbe words for under \$0.10, scalable to substantial corpora for under \$10.

\section{Conclusion and Future Work}
\label{sec:conclusion}

We presented an exploratory evaluation of six LLM elicitation strategies for extracting usable text data for Hausa and Fongbe. While the scale of this study is limited, our initial findings suggest three trends worth investigating further. First, GPT-4o Mini produces substantially more usable text than Gemini 2.5 Flash, yielding 6$\times$ more Hausa words and 41$\times$ more Fongbe words per API call. Second, elicitation strategies appear to be language-dependent: Hausa benefits from volume-maximizing tasks (functional text, dialogue), while Fongbe appears to require constraint-heavy prompts (constrained generation). Third, the Hausa--Fongbe performance gap is consistent across conditions, suggesting that LLM-based extraction may currently be more viable for moderately resourced languages. These findings are preliminary and will require validation at larger scale and across additional languages and models.


\label{sec:future}
Future work will include additional LLMs (Claude Sonnet, open-source and African-language-focused models), human evaluation with native speakers, downstream utility testing on MasakhaNER~2.0 and MasakhaPOS, target-language prompting experiments, larger prompt samples for statistical robustness, and extension to additional African languages.

\section{Limitations}
\label{sec:limitations}

This study has several limitations. First, we evaluate only two commercial LLMs; the performance gap we observe may not generalize to open-source or African-language-focused models. Second, our evaluation relies entirely on automatic metrics; human evaluation by native speakers is essential, particularly for Fongbe where GlotLID misidentifies 47\% of GPT-4o Mini outputs despite many likely containing valid Fongbe with code-switching --- misidentification does not necessarily imply low linguistic quality, but may reflect code-switching that the classifier penalizes. Third, we do not evaluate downstream task utility: whether extracted corpora improve NER or POS tagging performance remains to be tested. Fourth, with 25 prompts per task type, sample sizes are modest; while directional patterns are consistent across conditions, larger experiments would enable more robust statistical comparisons. Fifth, all prompts are written in English; target-language prompting may yield different results. Sixth, our memorization assessment relies on reference overlap as an indirect proxy, though the uniformly low cosine similarity values ($<$0.12) suggest verbatim reproduction is unlikely. Finally, our methodology relies on commercial APIs, introducing cost barriers and reproducibility concerns; future work should investigate open-source alternatives.

\section{Ethics Statement}
\label{sec:ethics}

\textbf{Data provenance and community benefit.} We acknowledge that LLMs were trained on data contributed by language communities, often without explicit consent. Our work aims to redirect encoded knowledge back to these communities. All generated data will be released under CC-BY-4.0.

\textbf{Quality and potential harms.} LLM-generated text may contain errors, inaccuracies, or hallucinated content. We document the synthetic nature of all corpora and recommend native speaker validation before production use.

\textbf{Commercial API usage.} Our methodology relies on commercial APIs, introducing cost barriers. Future work should investigate open-source alternatives.

\section{Data and Code Availability}

All generated corpora, prompts, generation scripts, and evaluation code will be made publicly available upon acceptance under a CC-BY-4.0 license. The evaluation pipeline, including the GlotLID-based language fidelity assessment, is provided for full reproducibility.



\section{Acknowledgements}
The authors thank the reviewers for their constructive feedback. We acknowledge the Masakhane community for their foundational contributions to African NLP resources, particularly the MasakhaNER~2.0 and MasakhaPOS datasets used in our evaluation. This publication was developed as part of the Center for Inclusive Digital Transformation of Africa (CIDTA) and the Afretec Network, which is managed by Carnegie Mellon University Africa and receives financial support from the Mastercard Foundation. The views expressed in this document are solely those of the authors and do not necessarily reflect those of Carnegie Mellon University or the Mastercard Foundation.

\section{Bibliographical References}
\label{sec:reference}

\bibliographystyle{lrec2026-natbib}
\bibliography{rail}

@inproceedings{Nekoto-etal-2020,
    title = {Participatory Research for Low-resourced Machine Translation: {A} Case Study in {A}frican Languages},
    author = {Nekoto, Wilhelmina and Marivate, Vukosi and Matsila, Tshinondiwa and Fasubaa, Timi and others},
    booktitle = {Findings of the Association for Computational Linguistics: EMNLP 2020},
    year = {2020},
    publisher = {Association for Computational Linguistics},
    pages = {2144--2160},
}

@inproceedings{Adelani-etal-2022-masakhaner,
    title = {{M}asakha{NER} 2.0: {A}frica-centric Transfer Learning for Named Entity Recognition},
    author = {Adelani, David and Neubig, Graham and Ruder, Sebastian and Rijhwani, Shruti and Beukman, Michael and Palen-Michel, Chester and Lignos, Constantine and Alabi, Jesujoba and Muhammad, Shamsuddeen and Nabende, Peter and Dione, Cheikh and others},
    booktitle = {Proceedings of the 2022 Conference on Empirical Methods in Natural Language Processing},
    year = {2022},
    publisher = {Association for Computational Linguistics},
}

@inproceedings{Adelani-etal-2022-mafand,
    title = {A Few Thousand Translations Go a Long Way! {L}everaging Pre-trained Models for {A}frican News Translation},
    author = {Adelani, David and Alabi, Jesujoba and Fan, Angela and Kreutzer, Julia and others},
    booktitle = {Proceedings of the 2022 Conference of the North American Chapter of the Association for Computational Linguistics},
    year = {2022},
    publisher = {Association for Computational Linguistics},
}

@inproceedings{Adelani-etal-2023-masakhpos,
    title = {{M}asakha{POS}: {P}art-of-Speech Tagging for Typologically Diverse {A}frican Languages},
    author = {Dione, Cheikh M Bamba and Adelani, David and others},
    booktitle = {Proceedings of the 61st Annual Meeting of the Association for Computational Linguistics},
    year = {2023},
    publisher = {Association for Computational Linguistics},
}

@misc{Adelani-etal-2024-fikira,
    title = {Fikira: Multilingual Reasoning Dataset for {A}frican Languages},
    author = {Adelani, David and Muhammad, Shamsuddeen and others},
    year = {2024},
    note = {Masakhane Project Technical Report},
}

@inproceedings{Alabi-etal-2025-afridocmt,
    title = {{AFRIDOC}-{MT}: Document-level {MT} Corpus for {A}frican Languages},
    author = {Alabi, Jesujoba Oluwadara and Azime, Israel Abebe and others},
    booktitle = {Proceedings of the 2025 Conference on Empirical Methods in Natural Language Processing},
    year = {2025},
    publisher = {Association for Computational Linguistics},
}

@inproceedings{Ogueji-etal-2021,
    title = {Small Data? {N}o Problem! {E}xploring the Viability of Pretrained Multilingual Language Models for Low-resourced Languages},
    author = {Ogueji, Kelechi and Zhu, Yuxin and Lin, Jimmy},
    booktitle = {Proceedings of the 1st Workshop on Multilingual Representation Learning},
    year = {2021},
    publisher = {Association for Computational Linguistics},
}

@inproceedings{Alabi-etal-2022,
    title = {Adapting Pre-trained Language Models to {A}frican Languages via Multilingual Adaptive Fine-Tuning},
    author = {Alabi, Jesujoba and Adelani, David and Mosbach, Marius and Klakow, Dietrich},
    booktitle = {Proceedings of the 29th International Conference on Computational Linguistics},
    year = {2022},
}

@inproceedings{Joshi-etal-2020,
    title = {The State and Fate of Linguistic Diversity and Inclusion in the {NLP} World},
    author = {Joshi, Pratik and Santy, Sebastin and Buber, Amar and Bali, Kalika and Choudhury, Monojit},
    booktitle = {Proceedings of the 58th Annual Meeting of the Association for Computational Linguistics},
    year = {2020},
    publisher = {Association for Computational Linguistics},
    pages = {6282--6293},
}

@inproceedings{Hedderich-etal-2021,
    title = {A Survey on Recent Approaches for Natural Language Processing in Low-Resource Scenarios},
    author = {Hedderich, Michael and Lange, Lukas and Adel, Heike and Strobe, Jannik and Klakow, Dietrich},
    booktitle = {Proceedings of the 2021 Conference of the North American Chapter of the Association for Computational Linguistics},
    year = {2021},
    publisher = {Association for Computational Linguistics},
}

@article{Orife-etal-2020,
    title = {Masakhane -- Machine Translation for {A}frica},
    author = {Orife, Iroro and Kreutzer, Julia and Dossou, Bonaventure and Emezue, Chris and others},
    journal = {arXiv preprint arXiv:2003.11529},
    year = {2020},
}

@inproceedings{Robinson-etal-2023,
    title = {{C}hat{GPT} {MT}: Competitive for High- (but not Low-) Resource Languages},
    author = {Robinson, Nathaniel and Ogayo, Perez and Mortensen, David R. and Neubig, Graham},
    booktitle = {Proceedings of the Eighth Conference on Machine Translation},
    year = {2023},
    publisher = {Association for Computational Linguistics},
}

@article{Hendy-etal-2023,
    title = {How Good Are {GPT} Models at Machine Translation? {A} Comprehensive Evaluation},
    author = {Hendy, Amr and Abdelrehim, Mohamed and Sharaf, Amr and Rauber, Vikas and others},
    journal = {arXiv preprint arXiv:2302.09210},
    year = {2023},
}

@inproceedings{Whitehouse-etal-2023,
    title = {{LLM}-powered Data Augmentation for Enhanced Cross-lingual Performance},
    author = {Whitehouse, Chenxi and others},
    booktitle = {Proceedings of the 2023 Conference on Empirical Methods in Natural Language Processing},
    year = {2023},
    publisher = {Association for Computational Linguistics},
}

@inproceedings{Schick-Schuetze-2021,
    title = {Generating Datasets with Pretrained Language Models},
    author = {Schick, Timo and Sch{\"u}tze, Hinrich},
    booktitle = {Proceedings of the 2021 Conference on Empirical Methods in Natural Language Processing},
    year = {2021},
    publisher = {Association for Computational Linguistics},
}

@inproceedings{Dai-Adel-2020,
    title = {An Analysis of Simple Data Augmentation for Named Entity Recognition},
    author = {Dai, Xiang and Adel, Heike},
    booktitle = {Proceedings of the 28th International Conference on Computational Linguistics},
    year = {2020},
}

@inproceedings{Wei-Zou-2019,
    title = {{EDA}: Easy Data Augmentation Techniques for Boosting Performance on Text Classification Tasks},
    author = {Wei, Jason and Zou, Kai},
    booktitle = {Proceedings of the 2019 Conference on Empirical Methods in Natural Language Processing},
    year = {2019},
    publisher = {Association for Computational Linguistics},
}

@inproceedings{Sennrich-etal-2016,
    title = {Improving Neural Machine Translation Models with Monolingual Data},
    author = {Sennrich, Rico and Haddow, Barry and Birch, Alexandra},
    booktitle = {Proceedings of the 54th Annual Meeting of the Association for Computational Linguistics},
    year = {2016},
    publisher = {Association for Computational Linguistics},
}

@inproceedings{Conneau-etal-2020,
    title = {Unsupervised Cross-lingual Representation Learning at Scale},
    author = {Conneau, Alexis and Khandelwal, Kartikay and Goyal, Naman and Chaudhary, Vishrav and Wenzek, Guillaume and Guzm{\'a}n, Francisco and Grave, Edouard and Ott, Myle and Zettlemoyer, Luke and Stoyanov, Veselin},
    booktitle = {Proceedings of the 58th Annual Meeting of the Association for Computational Linguistics},
    year = {2020},
    publisher = {Association for Computational Linguistics},
    pages = {8440--8451},
}

@book{Lefebvre-Brousseau-2002,
    title = {A Grammar of {F}ongbe},
    author = {Lefebvre, Claire and Brousseau, Anne-Marie},
    year = {2002},
    publisher = {Mouton de Gruyter},
    address = {Berlin},
}

@book{Newman-2000,
    title = {The {H}ausa Language: An Encyclopedic Reference Grammar},
    author = {Newman, Paul},
    year = {2000},
    publisher = {Yale University Press},
}

@misc{Kargaran-etal-2023,
    title = {{GlotLID}: Language Identification for Low-Resource Languages},
    author = {Kargaran, Amir Hossein and Imani, Ayyoob and Yvon, Fran{\c{c}}ois and Sch{\"u}tze, Hinrich},
    year = {2023},
    howpublished = {\url{https://huggingface.co/cis-lmu/glotlid}},
    note = {Version 1.0},
}




\appendix

\renewcommand{\thesection}{Appendix \Alph{section}}

\section{Code and Data Repository}
\label{app:repo}

All prompts, generation scripts, evaluation code, and generated corpora are publicly available at:

\url{https://github.com/Pericles001/mining_llm_low_resource_languages_fon_hau/tree/main}

The repository is organised as follows:

\begin{description}
    \item[\texttt{prompts/}] JSON files containing all 150 prompts per language, organised by task type
    \item[\texttt{src/}] Core modules for generation (\texttt{generator.py}), evaluation (\texttt{evaluator.py}), and language detection (\texttt{language\_detector.py})
    \item[\texttt{scripts/}] CLI entry points for generation, evaluation, and analysis
    \item[\texttt{outputs/}] Raw LLM outputs organised by model, language, and task type
    \item[\texttt{results/}] Aggregated evaluation results, figures, and \LaTeX{} tables
\end{description}


\section{Supplementary Figures}
\label{app:figures}

\begin{figure}[H]
\centering
\includegraphics[width=0.45\textwidth]{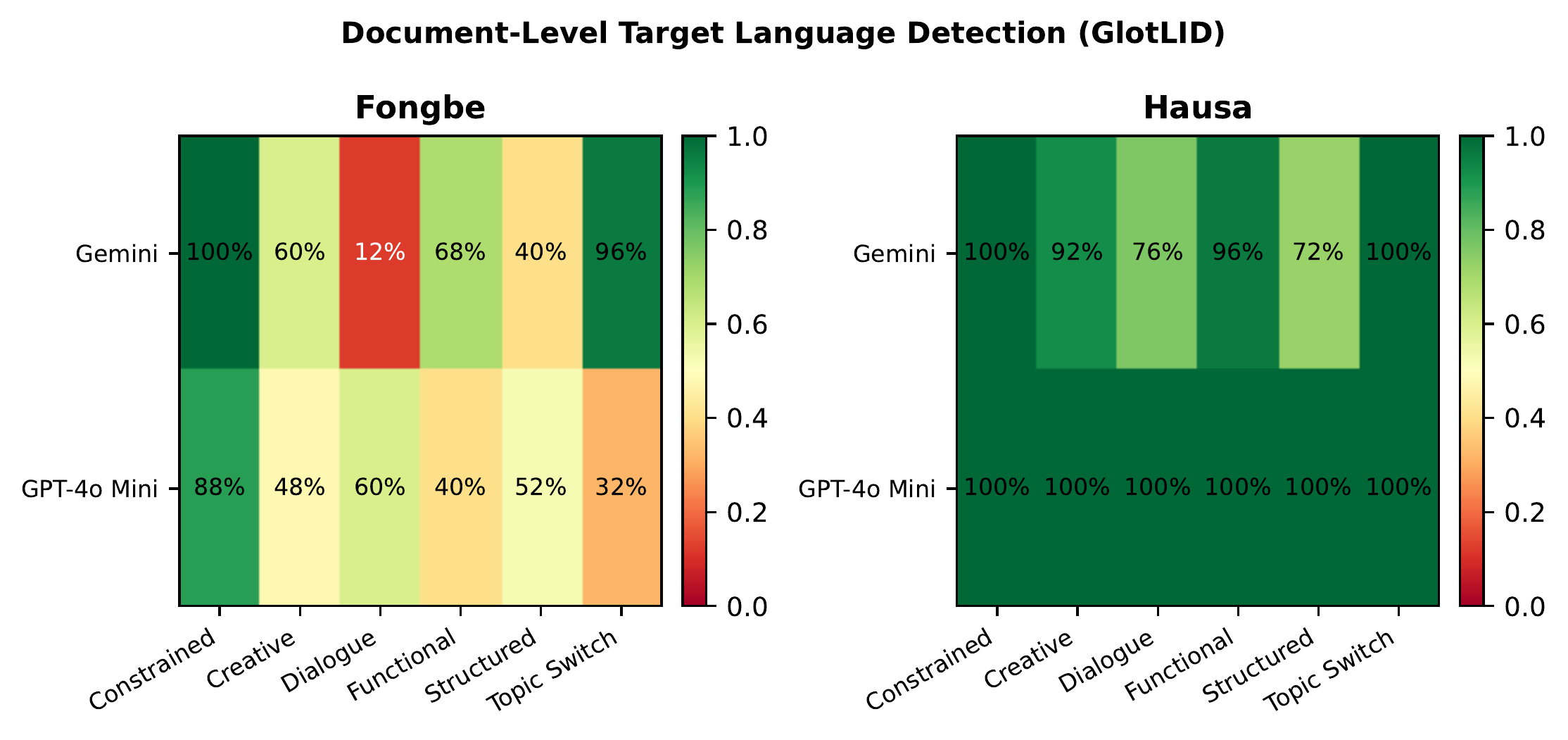}
\caption{Document-level target language detection heatmap (GlotLID). Green = high fidelity; red = low. Hausa is uniformly high for GPT-4o Mini; Fongbe fidelity depends strongly on task type.}
\label{fig:fidelity}
\end{figure}

\begin{figure}[H]
\centering
\includegraphics[width=0.45\textwidth]{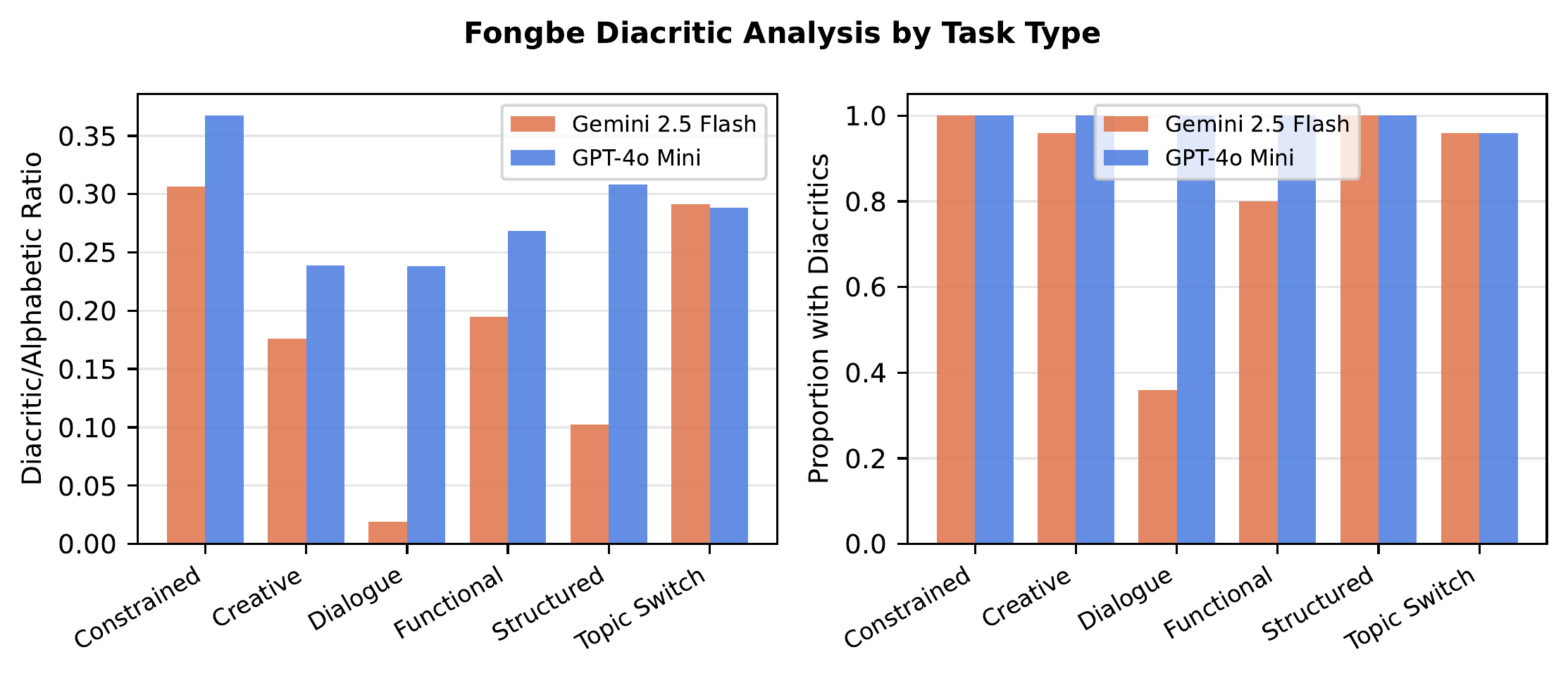}
\caption{Fongbe diacritic analysis by task type. Left: diacritic-to-alphabetic ratio; Right: proportion of outputs containing any diacritics. Constrained generation reliably elicits diacritics from both models.}
\label{fig:diacritics}
\end{figure}

\begin{figure}[H]
\centering
\includegraphics[width=0.45\textwidth]{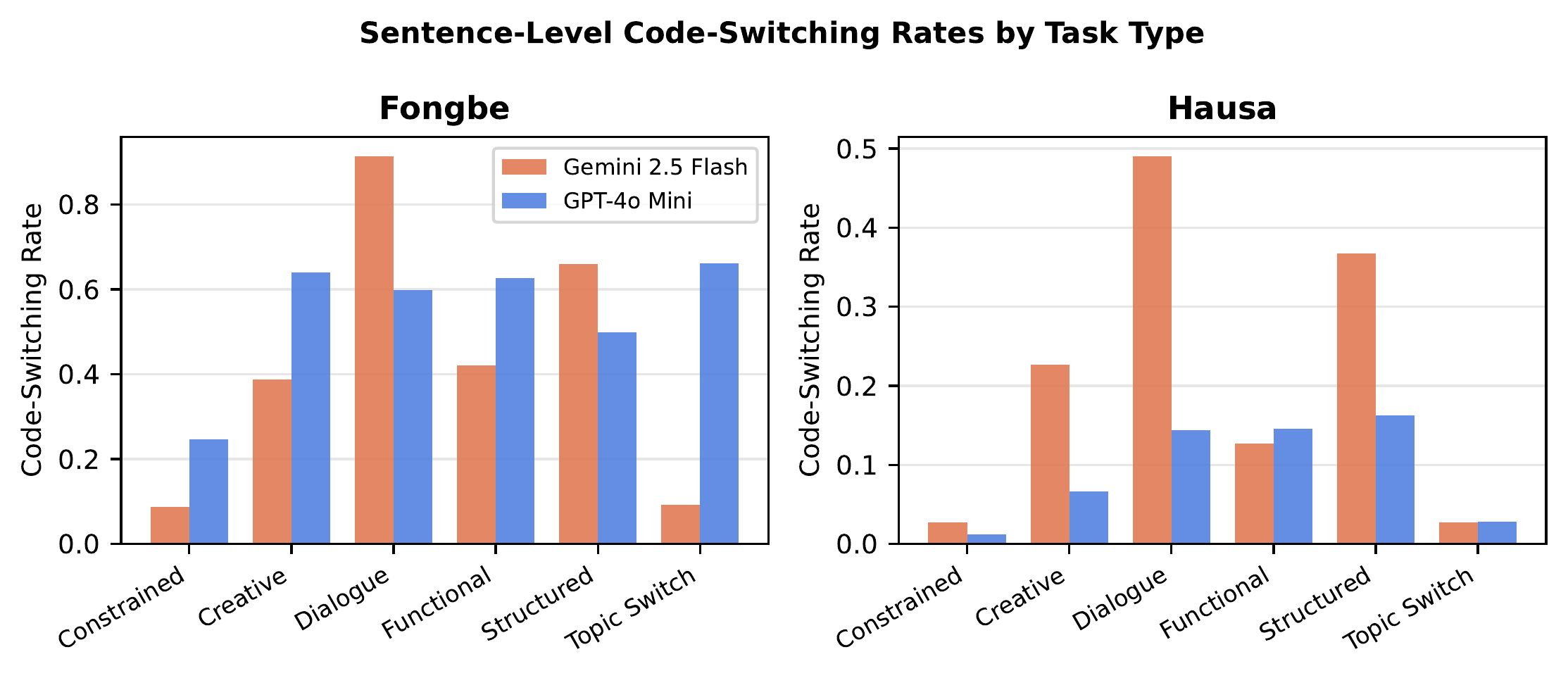}
\caption{Sentence-level code-switching rates by model, language, and task type. Constrained generation consistently achieves the lowest code-switching. Fongbe shows much higher rates than Hausa across all tasks.}
\label{fig:codeswitching}
\end{figure}

\section{Full Evaluation Summary}
\label{app:summary}

Table~\ref{tab:full_summary} reports all evaluation metrics across all 24 conditions (2 models $\times$ 2 languages $\times$ 6 task types). \textit{Quality} is a composite score averaging language confidence and inverse code-switching rate.

\begin{table*}[h]
\centering
\small
\setlength{\tabcolsep}{4pt}
\begin{tabular}{llllrrrrrrrr}
\toprule
\textbf{Model} & \textbf{Lang} & \textbf{Task} & \textbf{Valid\%} & \textbf{Words} & \textbf{TTR} & \textbf{Hapax} & \textbf{Vocab} & \textbf{CS} & \textbf{LangConf} & \textbf{Quality} \\
\midrule
\multirow{12}{*}{Gemini} & \multirow{6}{*}{fon}
  & constrained & 0.20 & 17.1 & 0.891 & 0.802 & 14.7 & 0.087 & 0.998 & 0.927 \\
 & & creative     & 0.28 & 17.6 & 0.932 & 0.876 & 16.4 & 0.387 & 0.782 & 0.791 \\
 & & dialogue     & 0.80 & 22.8 & 0.880 & 0.787 & 20.2 & 0.913 & 0.744 & 0.555 \\
 & & functional   & 0.36 & 16.7 & 0.883 & 0.795 & 14.7 & 0.420 & 0.929 & 0.816 \\
 & & structured   & 0.40 & 18.5 & 0.955 & 0.915 & 17.7 & 0.660 & 0.616 & 0.645 \\
 & & topic switch & 0.04 & 15.0 & 0.892 & 0.800 & 13.4 & 0.093 & 0.995 & 0.941 \\
\cline{2-11}
 & \multirow{6}{*}{hau}
  & constrained & 0.92 & 34.0 & 0.822 & 0.704 & 27.2 & 0.027 & 1.000 & 0.921 \\
 & & creative     & 0.76 & 26.8 & 0.918 & 0.854 & 23.5 & 0.227 & 0.912 & 0.867 \\
 & & dialogue     & 0.52 & 19.9 & 0.920 & 0.859 & 18.2 & 0.490 & 0.873 & 0.817 \\
 & & functional   & 0.68 & 20.1 & 0.923 & 0.853 & 18.5 & 0.127 & 0.995 & 0.914 \\
 & & structured   & 0.56 & 20.1 & 0.946 & 0.904 & 18.9 & 0.367 & 0.856 & 0.779 \\
 & & topic switch & 0.88 & 72.8 & 0.812 & 0.707 & 52.7 & 0.027 & 1.000 & 0.919 \\
\midrule
\multirow{12}{*}{GPT-4o} & \multirow{6}{*}{fon}
  & constrained & 1.00 & 77.6  & 0.544 & 0.375 & 31.7 & 0.246 & 0.937 & 0.869 \\
 & & creative     & 1.00 & 90.0  & 0.581 & 0.405 & 50.0 & 0.640 & 0.703 & 0.688 \\
 & & dialogue     & 1.00 & 144.5 & 0.458 & 0.275 & 65.5 & 0.598 & 0.868 & 0.736 \\
 & & functional   & 1.00 & 153.0 & 0.479 & 0.325 & 73.5 & 0.627 & 0.870 & 0.675 \\
 & & structured   & 1.00 & 91.8  & 0.597 & 0.486 & 48.1 & 0.498 & 0.862 & 0.731 \\
 & & topic switch & 1.00 & 156.8 & 0.477 & 0.321 & 70.4 & 0.661 & 0.828 & 0.646 \\
\cline{2-11}
 & \multirow{6}{*}{hau}
  & constrained & 1.00 & 124.6 & 0.667 & 0.520 & 73.6 & 0.012 & 1.000 & 0.898 \\
 & & creative     & 1.00 & 103.7 & 0.674 & 0.512 & 67.5 & 0.066 & 1.000 & 0.887 \\
 & & dialogue     & 1.00 & 190.0 & 0.578 & 0.408 & 107.6 & 0.144 & 1.000 & 0.871 \\
 & & functional   & 1.00 & 204.8 & 0.602 & 0.448 & 117.4 & 0.146 & 1.000 & 0.874 \\
 & & structured   & 1.00 & 114.5 & 0.708 & 0.603 & 73.1 & 0.163 & 0.930 & 0.889 \\
 & & topic switch & 1.00 & 190.1 & 0.628 & 0.489 & 116.6 & 0.028 & 1.000 & 0.891 \\
\bottomrule
\end{tabular}
\caption{Full evaluation summary across all 24 conditions. CS = code-switching rate; LangConf = GlotLID language confidence score; Quality = composite score.}
\label{tab:full_summary}
\end{table*}

\clearpage
\section{Prompt Taxonomy Details}
\label{app:prompts}

This appendix documents the structure and rationale of all 150 prompts per language (6 task types $\times$ 25 prompts). Each task type is divided into subtasks to ensure domain coverage. All prompts use three placeholders: \texttt{\{language\}}, \texttt{\{language\_culture\}}, and \texttt{\{colonial\_language\}}, substituted at generation time.

\subsection*{A. Constrained Generation (cg\_01--cg\_25)}

Subtasks: \textit{vocabulary-constrained} (cg\_01--05), \textit{no-code-switching} (cg\_06--10), \textit{length-constrained} (cg\_11--15), \textit{technical-monolingual} (cg\_16--20), \textit{structure-constrained} (cg\_21--25).

\noindent\textbf{Design rationale:} Constrained generation prompts impose explicit linguistic constraints to prevent code-switching and test the model's ability to generate monolingual output. Vocabulary-constrained prompts seed the output with target-language words, reducing the risk of the model falling back to colonial language vocabulary for unknown concepts. Technical-monolingual prompts specifically target domains (computing, electricity, banking) where Fongbe and Hausa lack standard terminology, forcing the model to paraphrase rather than borrow.

\noindent\textbf{Representative templates:}
\begin{itemize}\small
    \item \texttt{cg\_01}: ``Write a short paragraph in \{language\} using ALL of the following words: \{word\_list\_1\}. Do not use any \{colonial\_language\} words.''
    \item \texttt{cg\_06}: ``Write a story in \{language\} about a day at the market. You must write ONLY in \{language\}. If you do not know a word in \{language\}, describe the concept using other \{language\} words instead of switching to \{colonial\_language\}.''
\end{itemize}

\noindent Word lists for vocabulary-constrained prompts are provided in the released data.

\subsection*{B. Creative Writing (cw\_01--cw\_25)}

Subtasks: \textit{poem} (cw\_01--05), \textit{folktale} (cw\_06--10), \textit{story} (cw\_11--15), \textit{song} (cw\_16--20), \textit{proverb} (cw\_21--25).

\noindent\textbf{Design rationale:} Creative writing prompts test deep cultural and linguistic knowledge by eliciting culturally rooted content (folktales, proverbs) that requires the model to draw on language-specific cultural knowledge, not just translation of English concepts. Folktales and proverbs are particularly valuable as they are community-specific and cannot easily be produced by back-translation.

\noindent\textbf{Representative templates:}
\begin{itemize}\small
    \item \texttt{cw\_06}: ``Write a traditional folktale in \{language\} about a clever tortoise who outsmarts a lion. The story should be 5--10 sentences long.''
\end{itemize}

\subsection*{C. Dialogue (dl\_01--dl\_25)}

Subtasks: \textit{conversation} (dl\_01--05), \textit{professional} (dl\_06--10), \textit{family} (dl\_11--15), \textit{interview} (dl\_16--20), \textit{negotiation} (dl\_21--25).

\noindent\textbf{Design rationale:} Dialogue prompts elicit colloquial register and spoken-form text, which is under-represented in formal corpora. The negotiation and professional subtasks target domains with specialized vocabulary (medical, agricultural, financial), which helps expand domain coverage of the resulting corpus.

\subsection*{D. Functional Text (ft\_01--ft\_25)}

Subtasks: \textit{letter} (ft\_01--05), \textit{instructions} (ft\_06--10), \textit{news} (ft\_11--15), \textit{recipe} (ft\_16--20), \textit{announcement} (ft\_21--25).

\noindent\textbf{Design rationale:} Functional text prompts target practical domains that are immediately useful for downstream NLP tasks (e.g., news classification, instruction following). These genres are typically well-represented in NLP benchmarks but under-resourced for African languages.

\subsection*{E. Structured Knowledge (sk\_01--sk\_25)}

Subtasks: \textit{definition} (sk\_01--05), \textit{cultural explanation} (sk\_06--10), \textit{grammar examples} (sk\_11--15), \textit{vocabulary list} (sk\_16--20), \textit{translation} (sk\_21--25).

\noindent\textbf{Design rationale:} Structured knowledge prompts elicit the model's metalinguistic knowledge, producing high-density lexical output (vocabulary lists, grammar examples) that is directly usable for dictionary construction and grammar documentation.

\subsection*{F. Topic Switching (ts\_01--ts\_25)}

Subtasks: \textit{domestic-to-sports} and related binary switches (ts\_01--10), \textit{narrative shift} (ts\_11--15), \textit{multi-topic} (ts\_16--20), \textit{knowledge switch} (ts\_21--25).

\noindent\textbf{Design rationale:} Topic-switching prompts stress-test language maintenance by requiring the model to continue in the target language after transitioning to a domain (technology, politics, science) that is more commonly discussed in the colonial contact language. This probes whether language fidelity holds under topic-induced pressure to code-switch.

\noindent\textbf{Representative templates:}
\begin{itemize}\small
    \item \texttt{ts\_25}: ``In \{language\}, describe a funeral ceremony. Then, in the same response and still in \{language\}, explain what artificial intelligence is.''
\end{itemize}

\end{document}